# Combining Multiple Time Series Models Through A Robust Weighted Mechanism

Ratnadip Adhikari
School of Computer and Systems Sciences
Jawaharlal Nehru University
New Delhi, India
Email: adhikari.ratan@gmail.com

R. K. Agrawal
School of Computer and Systems Sciences
Jawaharlal Nehru University
New Delhi, India
Email: rkajnu@gmail.com

*Abstract*—Improvement of time series forecasting accuracy through combining multiple models is an important as well as a dynamic area of research. As a result, various forecasts combination methods have been developed in literature. However, most of them are based on simple linear ensemble strategies and hence ignore the possible relationships between two or more participating models. In this paper, we propose a robust weighted nonlinear ensemble technique which considers the individual forecasts from different models as well as the correlations among them while combining. The proposed ensemble is constructed using three well-known forecasting models and is tested for three real-world time series. A comparison is made among the proposed scheme and three other widely used linear combination methods, in terms of the obtained forecast errors. This comparison shows that our ensemble scheme provides significantly lower forecast errors than each individual model as well as each of the four linear combination methods.

## I. INTRODUCTION

Time series forecasting is a continuously growing research area with fundamental importance in many domains of business, finance, demography, science and engineering, etc. Improvement of forecasting accuracies has received extensive attentions from researchers during the past two decades [1, 2]. In literature, it has been frequently observed that a suitable combination of multiple forecasts substantially improves the overall accuracy as well as outperforms each individual model [3, 4]. Combination strategies are the best intuitive alternatives to use, when there is a considerable amount of uncertainty associated with the selection of the optimal forecasting model. Moreover, combining multiple forecasts reduces the errors arising from faulty assumptions, bias, or mistakes in the data to a great extent.

Starting with the benchmark work of Bates and Granger in 1969 [5], a number of forecasts combination techniques have been developed in literature [6–8]. The most popular among them are the weighted linear combinations, where the weights assigned to the individual models are either equal or decided according to some rigorous mathematical rule. Some common linear forecasts combination methods are the simple average, trimmed average, Winsorized average, median, error-based method, outperformance method, variance-based pooling, etc. [9, 10]. Although linear combination techniques are easy to understand and implement, but they completely ignore the possible relationships among the participating models in the ensemble. This limitation has a negative effect on the forecasting accuracy of the ensemble, especially when the constituent forecasts are significantly correlated.

Nonlinear forecast combination is an area in which the existing literature is quite limited and so has a strong need of further developments [10]. In this paper, we propose a weighted nonlinear mechanism for combining forecasts from multiple time series models. Our proposed technique is partially motivated by the work of Freitas and Rodrigues [11] and it considers individual forecasts as well as correlations between pairs of forecasts for combining. Three forecasting models, viz. Autoregressive Integrated Moving Average (ARIMA), Artificial Neural Network (ANN) and Support Vector Machine (SVM) are used to build up the proposed ensemble. The appropriate combination weights are determined from the performances of the individual models on the validation datasets. The effectiveness of our proposed technique is empirically tested on three real-world time series, in terms of three common error measures: the Mean Absolute Percentage Error (MAPE), the Mean Squared Error (MSE), and the Average Relative Variance (ARV). Also, the forecasting performances of the proposed ensemble scheme for all three datasets are compared with three popular linear combination methods. These are the simple average, the median and a weighted linear combination in which the weights are determined from the validation MAPE values, obtained by the individual forecasting models.

The rest of the paper is organized as follows. Section II presents some well-known linear forecasts combination techniques. Our nonlinear ensemble scheme is described in Section III. In Section IV, we discuss about the three time series forecasting models, which are used here to build up the proposed ensemble. Experimental results are reported in Section V and in Section VI, we conclude this paper.

## II. LINEAR FORCASTS COMBINATION TECHNIQUES

In a linear combination technique, the combined forecast for the associated time series is calculated through a linear function of the individual forecasts from the contributing models. Let, $\mathbf{Y}=[y_1, y_2, \ldots, y_N]^\mathrm{T}$ be the actual time series, which is to be forecasted using $n$ different models and $\hat{\mathbf{Y}}^{(i)} = \left[\hat{y}_1^{(i)}, \hat{y}_2^{(i)}, \ldots, \hat{y}_N^{(i)}\right]^\mathrm{T}$ be its forecast obtained from the $i^\mathrm{th}$ model ($i$=1, 2,…, $n$). Then, a linear combination of these $n$ forecasted series of the original time series produces $\hat{\mathbf{Y}}^{(c)} = \left[\hat{y}_1^{(c)}, \hat{y}_2^{(c)}, ..., \hat{y}_N^{(c)}\right]^T$, given by:

$$\hat{y}_k^{(c)} = f\left(\hat{y}_k^{(1)}, \hat{y}_k^{(2)}, ..., \hat{y}_k^{(n)}\right) \forall k = 1, 2, ..., N.$$

where, $f$ is some linear function of the individual forecasts $\hat{y}_k^{(i)}$ ($i=1, 2,..., n$; $k=1, 2,..., N$). Thus, we have:

$$\left.\begin{array}{l}\hat{y}_k^{(c)} = w_1\hat{y}_k^{(1)} + w_2\hat{y}_k^{(2)} + ... + w_n\hat{y}_k^{(n)} = \sum_{i=1}^n w_i\hat{y}_k^{(i)} \\ \forall k = 1, 2, ..., N.\end{array}\right\} \quad (1)$$

Here, $w_i$ is the weight assigned to the $i^{th}$ forecasting method. To ensure unbiasedness, it is often assumed that the weights add up to unity. Some widely used linear combination techniques are briefly described below:

- In the *simple average*, all models are assigned equal weights, i.e. $w_i=1/n$ ($i=1, 2,..., n$) [9, 10].
- In the *trimmed average*, individual forecasts are combined by a simple arithmetic mean, excluding the worst performing $k$% of the models. A trimming of 10%–30% is usually recommended [9, 10].
- In the *Winsorized average*, the $i$ smallest and $i$ largest forecasts are selected and respectively set as the $(i+1)^{th}$ smallest and $(i+1)^{th}$ largest forecasts [9].
- In the *median-based* combining, the combination function $f$ is the median of the individual forecasts. Median is sometimes preferred over simple average as it is less sensitive to extreme values [12, 13].
- In the *error-based* combining, the weight to each model is assigned to be the inverse of the past forecast error (e.g. MSE, MAE, MAPE, etc.) of the corresponding model [3, 10].
- In the *variance-based* method, the optimal weights are determined through the minimization of the total Sum of Squared Error (SSE) [7, 10].

### III. THE PROPOSED ENSEMBLE TECHNIQUE

As mentioned earlier, a major disadvantage of a linear combination technique is that it considers only the contributions of the individual models, but totally overlooks the possible relationships among them. As a result, there is a considerable reduction in the forecasting accuracy of a linear combination scheme, when two or more participating models in the ensemble are correlated. To overcome this limitation, our ensemble technique is developed as an extension of the usual linear combination in order to deal with the possible correlations between pairs of forecasts.

#### A. Mathematical Description

Here we describe our ensemble method for combining three forecasts, but it can be easily generalized. Let, the actual test dataset of a time series be $\mathbf{Y}=[y_1, y_2, ..., y_N]^T$ with $\hat{\mathbf{Y}}^{(i)} = \left[\hat{y}_1^{(i)}, \hat{y}_2^{(i)}, ..., \hat{y}_N^{(i)}\right]^T$ being its forecast obtained from the $i^{th}$ method ($i=1, 2, 3$). Let, $\mu^{(i)}$ and $\sigma^{(i)}$ be the mean and standard deviation of $\hat{\mathbf{Y}}^{(i)}$, respectively. Then the combined forecast of $\mathbf{Y}$ is defined as:

$$\left.\begin{array}{l}\hat{\mathbf{Y}}^{(c)} = \left[\hat{y}_1^{(c)}, \hat{y}_2^{(c)}, ..., \hat{y}_N^{(c)}\right]^T, \\ \hat{y}_k^{(c)} = w_0 + w_1\hat{y}_k^{(1)} + w_2\hat{y}_k^{(2)} + w_3\hat{y}_k^{(3)} \\ \quad + \theta_1\hat{v}_k^{(1)}\hat{v}_k^{(2)} + \theta_2\hat{v}_k^{(2)}\hat{v}_k^{(3)} + \theta_3\hat{v}_k^{(3)}\hat{v}_k^{(1)} \\ \hat{v}_k^{(i)} = \left(\hat{y}_k^{(i)} - \mu^{(i)}\right) / \left(\sigma^{(i)}\right)^2 \\ i = 1, 2, 3; k = 1, 2, ..., N\end{array}\right\} \quad (2)$$

In (2), the nonlinear terms in calculating $\hat{y}_k^{(c)}$ are included to take into account the correlation effects between pairs of forecasts. The optimal weights are to be determined by minimizing the forecast SSE, given by:

$$\text{SSE} = \sum_{k=1}^N \left(y_k - \hat{y}_k^{(c)}\right)^2 \quad (3)$$

Now, for optimization of the combination weights, we have:

$$\left.\begin{array}{l}(\partial/\partial w_i)(\text{SSE}) = 0 \\ (\partial/\partial \theta_j)(\text{SSE}) = 0 \\ (i = 0, 1, 2, 3; j = 1, 2, 3)\end{array}\right\} \quad (4)$$

After computations of the associated partial derivatives in (4) and subsequent mathematical simplifications, we can get the following system of linear equations:

$$\left.\begin{array}{l}\mathbf{Vw} + \mathbf{Z\theta} = \mathbf{b} \\ \mathbf{Z}^T\mathbf{w} + \mathbf{U\theta} = \mathbf{d}\end{array}\right\}. \quad (5)$$

where,

$$\left.\begin{array}{l}\mathbf{w} = [w_0, w_1, w_2, w_3]^T, \mathbf{\theta} = [\theta_1, \theta_2, \theta_3]^T \\ \mathbf{V} = \mathbf{F}^T\mathbf{F}, \mathbf{Z} = \mathbf{F}^T\mathbf{G}, \mathbf{U} = \mathbf{G}^T\mathbf{G}, \mathbf{b} = \mathbf{F}^T\mathbf{Y}, \mathbf{d} = \mathbf{G}^T\mathbf{Y} \\ \mathbf{F} = \begin{bmatrix} 1 & \hat{y}_1^{(1)} & \hat{y}_1^{(2)} & \hat{y}_1^{(3)} \\ \vdots & \vdots & \vdots & \vdots \\ 1 & \hat{y}_N^{(1)} & \hat{y}_N^{(2)} & \hat{y}_N^{(3)} \end{bmatrix}_{N \times 4} \\ \mathbf{G} = \begin{bmatrix} \hat{v}_1^{(1)}\hat{v}_1^{(2)} & \hat{v}_1^{(2)}\hat{v}_1^{(3)} & \hat{v}_1^{(3)}\hat{v}_1^{(1)} \\ \vdots & \vdots & \vdots \\ \hat{v}_N^{(1)}\hat{v}_N^{(2)} & \hat{v}_N^{(2)}\hat{v}_N^{(3)} & \hat{v}_N^{(3)}\hat{v}_N^{(1)} \end{bmatrix}_{N \times 3}\end{array}\right\}.$$

Now, after solving the matrix equations in (5), the required optimal weights are obtained as:

$$\left.\begin{array}{l}\mathbf{\theta}_{opt} = \left(\mathbf{U} - \mathbf{Z}^T\mathbf{V}^{-1}\mathbf{Z}\right)^{-1}\left(\mathbf{d} - \mathbf{Z}^T\mathbf{V}^{-1}\mathbf{b}\right) \\ \mathbf{w}_{opt} = \mathbf{V}^{-1}\left(\mathbf{b} - \mathbf{Z}\mathbf{\theta}_{opt}\right)\end{array}\right\} \quad (6)$$

The optimal weights exist if and only if all the matrix inverses involved in (6) exist.

*B. Determination of the Combination Weights*

To determine the optimal weights in our proposed ensemble technique, the knowledge of the forecast SSE is required, but it is unknown in advance. So, we divide the available observations of the associated time series into a pair of *training* and *validation* subsets. The size of the *validation* set is chosen to be equal to the size of the out-of-sample *testing* dataset. The individual forecasting models are then trained on the training set and the optimal combination weights are calculated by minimizing the validation SSE. This approach of weights determination is especially suitable for time series showing regular patterns (e.g. stationary or seasonal series).

*C. Our Ensemble Algorithm*

Let $\mathbf{Y}=[y_1, y_2, \ldots, y_N]^T$ be the available observations of a time series and $M_i$ ($i=1, 2,\ldots, n$) be the $n$ forecasting models to be combined. Then, the steps in our proposed ensemble algorithm are outlined below:

1. Divide the dataset $\mathbf{Y}$ into a pair of *training* and *validation* subsets $\mathbf{Y}_{\text{train}}$ and $\mathbf{Y}_{\text{validation}}$, respectively as follows:
$$\mathbf{Y}_{\text{train}} = [y_1, y_2, \ldots, y_\alpha]^T$$
$$\mathbf{Y}_{\text{validation}} = [y_{\alpha+1}, y_{\alpha+2}, \ldots, y_N]^T$$
where,
$\alpha$ = size of the training set
$N - \alpha$ = size of the validation set
= size of the testing set

2. Formulate the equation for the combined forecast, as shown in (2).

3. Train each forecasting model $M_i$ ($i=1, 2,\ldots, n$) on $\mathbf{Y}_{\text{train}}$ to forecast the $\mathbf{Y}_{\text{validation}}$ dataset.

4. From the minimization of validation SSE, determine the optimal combination weight vectors $\mathbf{w}_{\text{comb}}$ and $\mathbf{\theta}_{\text{comb}}$ by using (6).

5. Use $\mathbf{w}_{\text{comb}}$ and $\mathbf{\theta}_{\text{comb}}$ to compute the combined forecasts for predicting the testing set.

IV. THE THREE CONSTITUENT FORECASTING MODELS

To build up our proposed nonlinear ensemble, three popular time series forecasting models are used in this paper, which are: the Autoregressive Integrated Moving Average (ARIMA), the Support Vector Machine (SVM) and the Artificial Neural Network (ANN). Brief descriptions of these three models are presented below.

*A. The ARIMA Model*

The ARIMA models, developed by Box and Jenkins [2] are the most popular statistical methods for time series forecasting. An ARIMA($p$, $d$, $q$) model is given by:
$$\phi(L)(1-L)^d y_t = \theta(L)\varepsilon_t \tag{7}$$
where,
$$\phi(L) = 1 - \sum_{i=1}^{p}\phi_i L^i, \theta(L) = 1 + \sum_{j=1}^{q}\theta_j L^j \text{ and } Ly_t = y_{t-1}$$

Here, the model orders $p$, $d$, $q$ represent *autoregressive*, *degree of differencing* and *moving average* processes, respectively; $y_t$ is the actual time series and $\varepsilon_t$ is a white noise process. In this model, a nonstationary time series is transformed to a stationary one by successively ($d$ times) differencing it [2, 4]. For real world applications, a single differencing is often sufficient.

*B. The SVM Model*

During the past few years, SVMs have gained notable popularity in the forecasting domain. These are a class of robust statistical models, developed by Vapnik and co-workers in 1995 and are based on the Structural Risk Minimization (SRM) principle [14]. Time series forecasting is a branch of Support Vector Regression (SVR) in which the principal aim is to construct an optimal separating hyperplane to correctly classify real-valued outputs. Given a training dataset $\{\mathbf{x}_i, y_i\}_{i=1}^{N}$ with $\mathbf{x}_i \in \mathbb{R}^n$, $y_i \in \mathbb{R}$, the SVM method attempts to approximate the unknown data generation function in the form: $f(\mathbf{x})=\mathbf{w}\cdot\varphi(\mathbf{x})+b$, where $\mathbf{w}$ is the weight vector, $\varphi$ is the nonlinear mapping to a higher dimensional feature space and $b$ is the bias term.

The SVM regression is converted to a Quadratic Programming Problem (QPP), using Vapnik's ε-insensitive loss function [14, 15] in order to minimize the empirical risk. After solving the associated QPP, the optimal decision hyperplane is given by:
$$y(\mathbf{x}) = \sum_{i=1}^{N_s}(\alpha_i - \alpha_i^*)K(\mathbf{x}, \mathbf{x}_i) + b_{\text{opt}} \tag{8}$$

where, $\alpha_i, \alpha_i^*$ are the Lagrange multipliers ($i=1, 2,\ldots, N_s$), $K(\mathbf{x}, \mathbf{x}_i)$ is the kernel function, $N_s$ is the number of support vectors and $b_{\text{opt}}$ is the optimal bias. In this paper, the Radial Basis Function (RBF) kernel: $K(\mathbf{x}, \mathbf{y})=\exp(-\|\mathbf{x}-\mathbf{y}\|^2/2\sigma^2)$ is used and the proper SVM hyper parameters are selected through cross validation techniques.

*C. The ANN Model*

ANNs are a class of nonlinear, nonparametric, data-driven and self-adaptive models, originally inspired by the intelligent working mechanism of human brains [4, 16]. Over the years, ANNs are used as excellent alternative to the common statistical models for time series forecasting. The most popular ANN architectures in forecasting domain are the *Multilayer Perceptrons (MLPs)*. They consist of a feedforward structure of three layers, viz. an input layer, one or more hidden layer and an output layer. The nodes in each layer are connected to those in the immediate next layer by acyclic links [16]. Single hidden layer is sufficient for most applications. It is often customary to use the notation ($p$, $h$, $q$) for referring an ANN with $p$ input, $h$ hidden and $q$ output nodes. A typical MLP architecture is shown in Fig. 1.

The forecasting performance of an ANN model depends on a number of factors, e.g. the selection of proper network architecture, training algorithm, activation functions, significant time lags, etc [16, 17]. Unfortunately, no rigorous theoretical procedure is available in this regard and often these issues have to be resolved experimentally.

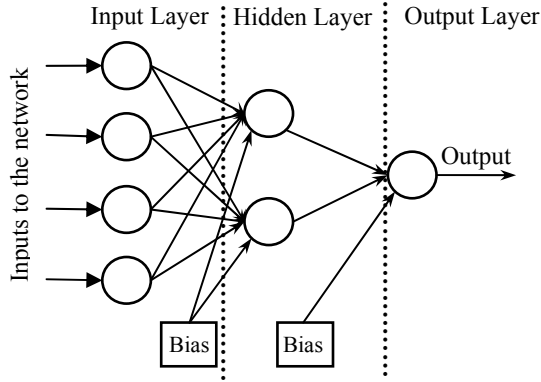

Figure 1. A typical MLP architecture

In this paper, we use the *Resilient Propagation (RP)* [18] as the network training algorithm and the logistic and identity functions as the hidden and output layer activation functions, respectively. Cross validation techniques are adopted, when necessary for selecting the appropriate ANN structure for a time series.

## V. EXPERIMENTAL RESULTS AND DISCUSSIONS

For empirical verification of forecasting performances of our proposed ensemble technique, three real world time series are used in this paper. These are the Canadian lynx, the Wolf's sunspots and the monthly international airline passengers series. All the three series are available on the well-known Time Series Data Library (TSDL) [19]. The description of these three time series is presented in Table I and their corresponding time plots are shown in Fig. 2.

TABLE I.    DESCRIPTION OF THE TIME SERIES DATASETS

| Time Series | Description | Size |
| --- | --- | --- |
| Lynx[a] | Number of lynx trapped per year in the Mackenzie River district of Northern Canada (1821–1934). | Total size: 114 Testing size: 14 |
| Sunspots | The annual number of observed sunspots (1700–1987). | Total size: 288 Testing size: 67 |
| Airline Passengers | Monthly number of international airline passengers (in thousands) (January 1949–December 1960). | Total size: 144 Testing size: 12 |

a. The logarithms (to the base 10) of the observations are used

The experiments in this paper are implemented on MATLAB; the neural network toolbox [20] is used for fitting the ANN models. Forecasting performances of all the models are evaluated in terms of three well-known error statistics, viz. the Mean Absolute Percentage Error (MAPE), the Mean Squared Error (MSE), and the Average Relative Variance (ARV). These are defined below:

$$\text{MAPE} = \frac{1}{N} \sum_{t=1}^{N} \left| \frac{y_t - \hat{y}_t}{y_t} \right| \times 100.$$

$$\text{MSE} = \frac{1}{N} \sum_{t=1}^{N} (y_t - \hat{y}_t)^2.$$

$$\text{ARV} = \left( \sum_{t=1}^{N} (y_t - \hat{y}_t)^2 \right) \bigg/ \left( \sum_{t=1}^{N} (\mu - \hat{y}_t)^2 \right).$$

Here, $y_t$ and $\hat{y}_t$ are the actual and forecasted observations, respectively; $N$ is the size and $\mu$ is the mean of the test set. The values of all these error measures are desired to be as less as possible for an efficient forecasting performance.

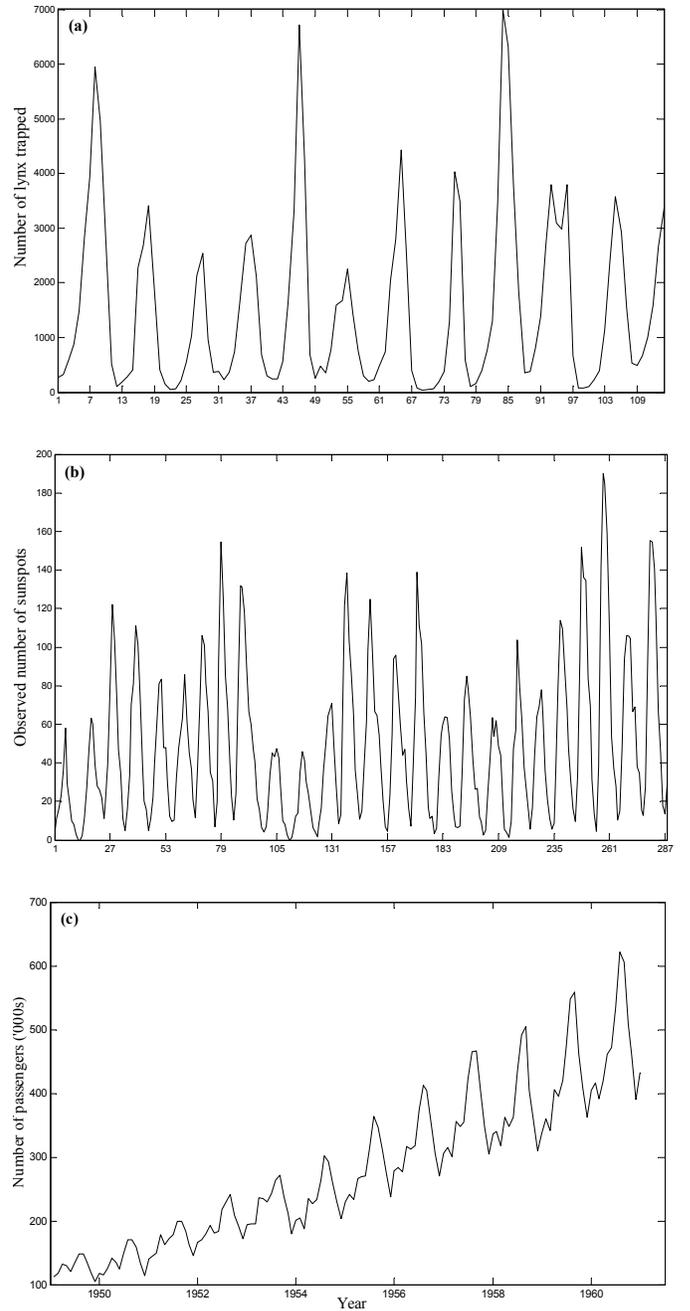

Figure 2. Time plots: (a) lynx, (b) sunspots, (c) airline passengers

The Canadian lynx and Wolf's sunspots are stationary time series, having approximate cycles of 10 and 11 years, respectively. Both these series have been extensively studied in literature. Following notable previous works [4, 22], in this paper we fit the ARIMA(12, 0, 0) (i.e. AR(12)) and a (7, 5, 1) ANN to the lynx series, while the ARIMA(9, 0, 0) (i.e. AR(9)) and a (4, 4, 1) ANN to the sunspots series.

The airline time series shows a multiplicative, monthly seasonal pattern with an upward trend. The most suitable statistical model for this series is the Seasonal ARIMA (SARIMA) of order $(0, 1, 1) \times (0, 1, 1)^{12}$, a variation of the basic ARIMA model [2, 17, 21]. This model is used in this paper for the airline data. For neural network modeling, the Seasonal ANN (SANN) structure, developed by Hamzacebi in 2008 [21] is employed. For a seasonal time series with period $s$, the SANN considers a $(s, h, s)$ ANN structure, $h$ being the number of hidden nodes. This model is quite simple to understand and apply, yet found to be very effective in modeling seasonal data [21]. In this paper, the (12, 1, 12) SANN is used for the airline time series.

Our proposed ensemble mechanism is compared with three other popular linear combination techniques. These are the simple average, the median and an error-based weighted linear combination in which the weight to an individual model is assigned to be the inverse of the corresponding validation MAPE value. The obtained forecasting results of all the fitted models for all three time series are depicted in Table II.

TABLE II. FORECSATING RESULTS FOR THE THREE TIME SERIES

| **Error Measures** | | **Lynx** | **Sunspots** | **Airline**[a] |
|---|---|---|---|---|
| **ARIMA** | MAPE | 3.277425 | 60.03853 | 3.709594 |
| | MSE | 0.012849 | 483.4907 | 0.041177 |
| | ARV | 0.070715 | 0.216361 | 0.076589 |
| **SVM** | MAPE | 5.811812 | 40.43313 | 2.336608 |
| | MSE | 0.052676 | 792.9613 | 0.017689 |
| | ARV | 0.279036 | 0.715230 | 0.029238 |
| **ANN** | MAPE | 2.912820 | 35.88591 | 2.577642 |
| | MSE | 0.013675 | 341.0395 | 0.019601 |
| | ARV | 0.098152 | 0.163557 | 0.033675 |
| **Simple Average** | MAPE | 2.807244 | 32.74193 | 2.373179 |
| | MSE | 0.013812 | 379.6589 | 0.019099 |
| | ARV | 0.086445 | 0.251363 | 0.033634 |
| **Median** | MAPE | 2.742802 | 33.42647 | 2.480301 |
| | MSE | 0.013004 | 331.5190 | 0.020184 |
| | ARV | 0.083436 | 0.206604 | 0.035741 |
| **Weighted Linear Comb.** | MAPE | 2.714213 | 31.84739 | 2.369855 |
| | MSE | 0.013142 | 311.0158 | 0.018612 |
| | ARV | 0.082196 | 0.170267 | 0.032665 |
| **Proposed Ensemble** | MAPE | 2.691642 | 30.01823 | 2.317835 |
| | MSE | 0.008523 | 275.7206 | 0.016017 |
| | ARV | 0.059014 | 0.149325 | 0.029592 |

a. Original MSE=Obtained MSE×$10^4$

The presented results in Table II show that our ensemble technique has provided lowest forecast errors among all fitted models for all the three time series. From these empirical findings, it is quite evident that significant improvement in forecasting accuracies can be achieved by employing the proposed nonlinear ensemble technique. However, it should be noted that this technique is suitable when the constituent forecasts are strongly correlated to each other.

In this paper, the term *Forecast Diagram* is used to refer the graph which shows the actual and forecasted values for a time series. The three forecast diagrams, obtained through our ensemble method are presented in Fig. 3.

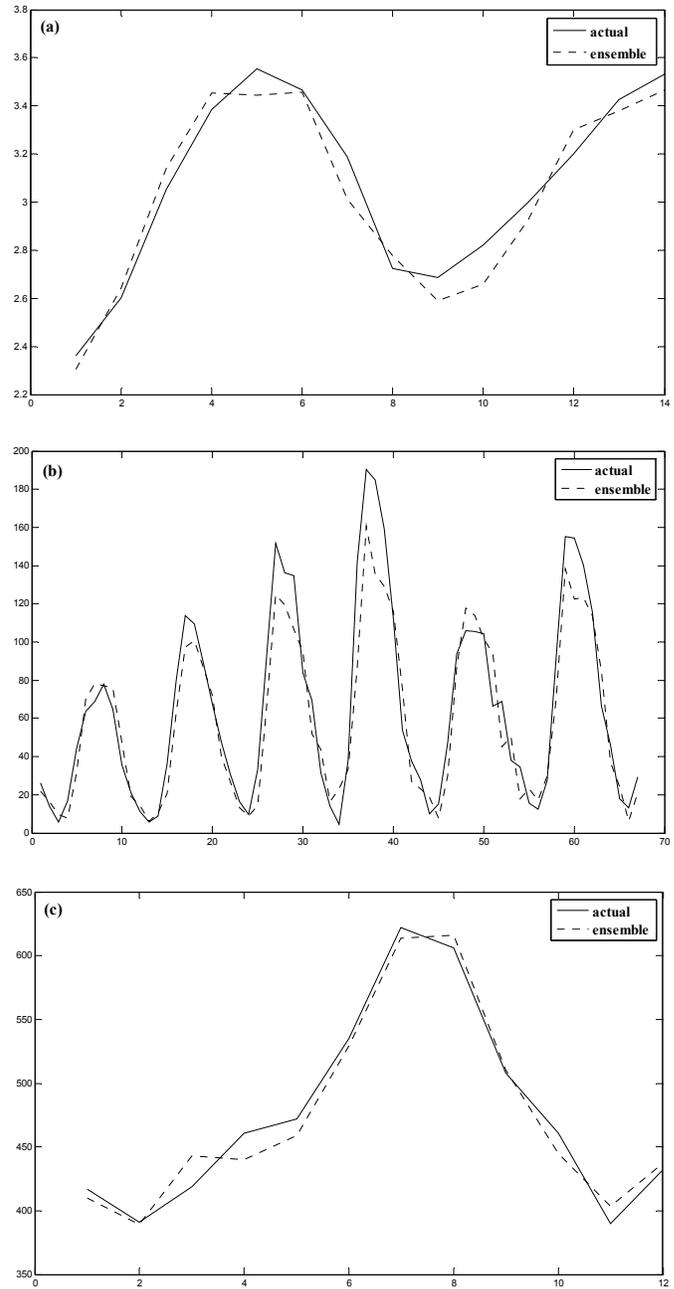

Figure 3. Forecast diagrams: (a) lynx, (b) sunspots, (c) airline passengers

## VI. Conclusions

Combining forecasts from conceptually different models effectively reduces the prediction errors and hence provides considerably increased accuracy. Over the years, many linear forecasts combination techniques have been developed in literature. Although these are simple to understand and implement but often criticized due to their ignorance of the relationships among contributing forecasts. Literature on nonlinear forecast combination is very limited and so there is a need of more extensive works on this area.

In this paper, we propose a robust weighted nonlinear technique for combining multiple time series models. The proposed method considers individual forecasts as well as the correlations between forecast pairs for creating the ensemble. A rigorous algorithm is suggested to determine the appropriate combination weights. The proposed ensemble is constructed with three well-known forecasting models and is tested on three real world time series, two stationary and one seasonal. Empirical findings demonstrate that the proposed technique outperforms each individual model as well as three other popular linear combination techniques, in terms of obtained forecast accuracies. It is to be noted that the proposed mechanism is highly efficient when the contributing forecasts are strongly correlated. In future, the effectiveness of the suggested method can be further explored, especially for nonstationary and chaotic time series datasets.

## Acknowledgment

The first author gratefully thanks the Council of Scientific and Industrial Research (CSIR) for the obtained financial assistance to carry out this research work.